\documentclass[runningheads]{llncs}
\usepackage{graphicx} 	
\usepackage{amsmath}
\usepackage[utf8]{inputenc} 
\usepackage[T1]{fontenc}    
\usepackage{hyperref}       
\usepackage{url}            
\usepackage{booktabs}       
\usepackage{amsfonts}       
\usepackage{nicefrac}       
\usepackage{microtype}      

\usepackage{amsmath}
\usepackage{graphicx}
\usepackage{xcolor}
\usepackage{etoolbox}
\usepackage[colorinlistoftodos]{todonotes}
\usepackage{makecell}
\usepackage{subcaption}

\begin{document}

\title{Learning Neural Models for End-to-End Clustering}

\author{Benjamin Bruno Meier\inst{1,2} \and
Ismail Elezi\inst{1,3} \and
Mohammadreza Amirian\inst{1,4} \and
\\Oliver D\"urr\inst{1,5} \and
Thilo Stadelmann\inst{1}}

\authorrunning{Meier, Elezi, Amirian, D\"urr \& Stadelmann}

\institute{ZHAW Datalab \& School of Engineering, Winterthur, Switzerland \and
ARGUS DATA INSIGHTS Schweiz AG, Zurich, Switzerland \and
Ca' Foscari University of Venice, Venice, Italy \and
Institute of Neural Information Processing, Ulm University, Germany \and
Institute for Optical Systems, HTWG Konstanz, Germany}

\maketitle

\begin{abstract}
We propose a novel end-to-end neural network architecture that, once trained, directly outputs a probabilistic clustering of a batch of input examples in one pass. It estimates a distribution over the number of clusters $k$, and for each $1 \leq k \leq k_\mathrm{max}$, a distribution over the individual cluster assignment for each data point. The network is trained in advance in a supervised fashion on separate data to learn grouping by any perceptual similarity criterion based on pairwise labels (same/different group). It can then be applied to different data containing different groups. We demonstrate promising performance on high-dimensional data like images (COIL-100) and speech (TIMIT). We call this ``learning to cluster'' and show its conceptual difference to deep metric learning, semi-supervise clustering and other related approaches while having the advantage of performing learnable clustering fully end-to-end. 
\end{abstract}

\setcounter{footnote}{0} 

\section{Introduction}
\begin{figure}[t]
  \centering
  \includegraphics[width=1.0\columnwidth]{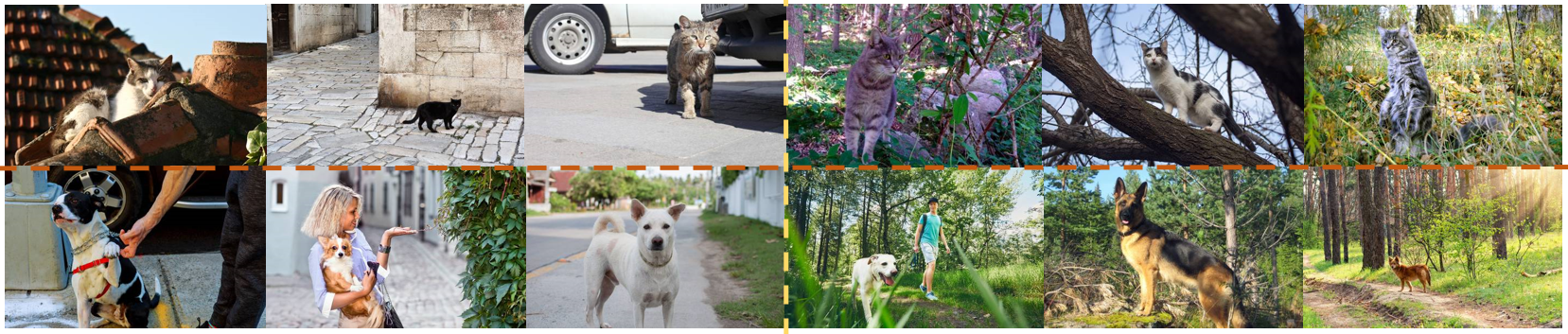}
  \caption{Images of cats (top) and dogs (bottom) in urban (left) and natural (right) environments.}   
  \label{fig:cats_dogs}
  \vspace{-0.5cm}
\end{figure}

Consider the illustrative task of grouping images of cats and dogs by \emph{perceived} similarity: depending on the intention of the user behind the task, the similarity could be defined by animal type (foreground object), environmental nativeness (background landscape, cp. Fig. \ref{fig:cats_dogs}) etc. This is characteristic of clustering perceptual, high-dimensional data like images \cite{kampffmeyer2017} or sound \cite{lukic2017speaker}: a user typically has some similarity criterion in mind when thinking about naturally arising groups (e.g., pictures by holiday destination, or persons appearing; songs by mood, or use of solo instrument). As defining such a similarity for every case is difficult, it is desirable to learn it. At the same time, the learned model will in many cases not be a classifier---the task will not be solved by classification---since the number and specific type of groups present at application time are not known in advance (e.g., speakers in TV recordings; persons in front of a surveillance camera; object types in the picture gallery of a large web shop). 

Grouping objects with machine learning is usually approached with clustering algorithms \cite{kaufman1990finding}. Typical ones like K-means \cite{macqueen1967some}, EM \cite{jin2011expectation}, hierarchical clustering \cite{murtagh1983survey} with chosen distance measure, or DBSCAN \cite{ester1996density} each have a specific inductive bias towards certain similarity structures present in the data (e.g., K-means: Euclidean distance from a central point; DBSCAN: common point density). Hence, to be applicable to above-mentioned tasks, they need high-level features that already encode the aspired similarity measure. This may be solved by learning salient embeddings \cite{mikolov2013efficient} with a deep metric learning approach \cite{hoffer2015deep}, followed by an off-line clustering phase using one of the above-mentioned algorithm.

However, it is desirable to combine these distinct phases (learning salient features, and subsequent clustering) into an end-to-end approach that can be trained globally \cite{lecun1998gradient}: it has the advantage of each phase being perfectly adjusted to the other by optimizing a global criterion, and removes the need of manually fitting parts of the pipeline. Numerous examples have demonstrated the success of neural networks for end-to-end approaches on such diverse tasks as speech recognition \cite{amodei2016deep}, robot control \cite{levine2016end}, scene text recognition \cite{shi2017end}, or music transcription \cite{sigtia2016end}.

In this paper, we present a conceptually novel approach that we call \emph{``learning to cluster''} in the above-mentioned sense of grouping high-dimensional data by some perceptually motivated similarity criterion. For this purpose, we define a novel neural network architecture with the following properties: (a) during training, it receives pairs of similar or dissimilar examples to learn the intended similarity function implicitly or explicitly; (b) during application, it is able to group objects of groups never encountered before; (c) it is trained end-to-end in a supervised way to produce a tailor-made clustering model and (d) is applied like a clustering algorithm to find both the number of clusters as well as the cluster membership of test-time objects in a fully probabilistic way.

\begin{figure}[t!]
  \centering
  \includegraphics[width=0.85\columnwidth]{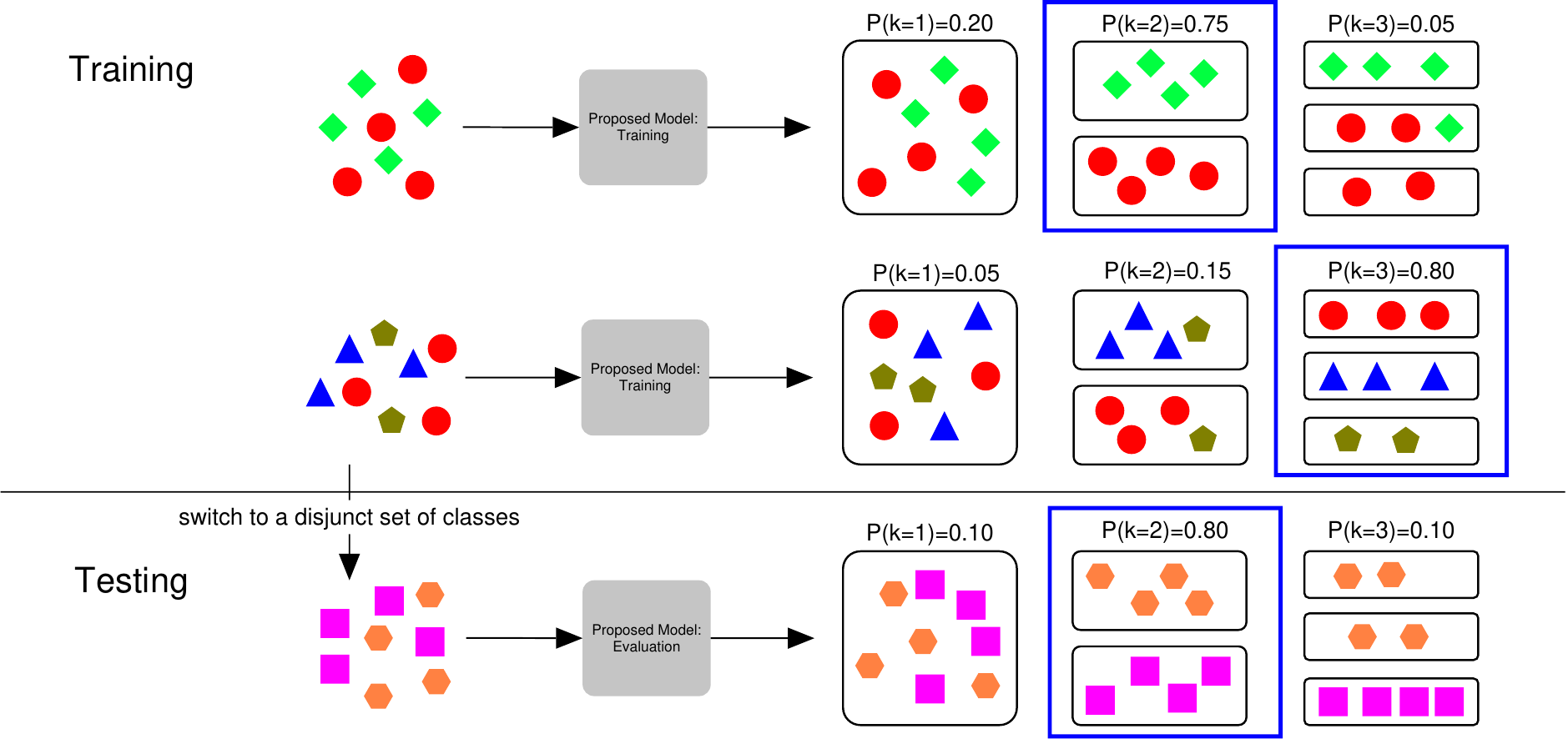}
  \caption{Training vs. testing: cluster types encountered during application/inference are never seen in training. Exemplary outputs (right-hand side) contain a partition for each $k$ ($1$--$3$ here) and a corresponding probability (best highlighted blue).}
  \label{fig:training_vs_evaluation}
  \vspace{-0.6cm}
\end{figure}

Our approach builds upon ideas from \emph{deep metric embedding}, namely to learn an embedding of the data into a representational space that allows for specific perceptual similarity evaluation via simple distance computation on feature vectors. However, it goes beyond this by adding the actual clustering step---grouping by similarity---directly to the same model, making it trainable end-to-end. Our approach is also different from \emph{semi-supervised clustering} \cite{Basu02semi-supervisedclustering}, which uses labels for some of the data points in the inference phase to guide the creation of groups. In contrast, our method uses absolutely no labels during inference, and moreover doesn't expect to have seen any of the groups it encounters during inference already during training (cp. Fig. \ref{fig:training_vs_evaluation}). Its training stage may be compared to creating K-means, DBSCAN etc. in the first place: it creates a specific clustering model, applicable to data with certain similarity structure, and once created/trained, the model performs ``unsupervised learning'' in the sense of finding groups. Finally, our approach differs from traditional cluster \emph{analysis} \cite{kaufman1990finding} in how the clustering algorithm is applied: instead of looking for patterns in the data in an unbiased and exploratory way, as is typically the case in unsupervised learning, our approach is geared towards the use case where users know perceptually what they are looking for, and can make this explicit using examples.  We then learn appropriate features and the similarity function simultaneously, taking full advantage of end-to-end learning. 

Our main contribution in this paper is the creation of a neural network architecture that learns to \emph{group} data, i.e., that outputs the same ``label'' for ``similar'' objects regardless of (a) it has ever seen this group before; (b) regardless of the actual value of the label (it is hence not a ``class''); and (c) regardless of the number of groups it will encounter during a single application run, up to a predefined maximum. This is novel in its concept and generality (i.e., learn to cluster previously unseen groups end-to-end for arbitrary, high-dimensional input without any optimization on test data). Due to this novelty in approach, we focus here on the general idea and experimental demonstration of the principal workings, and leave comprehensive hyperparameter studies and optimizations for future work. In Sec. \ref{sec:related_work}, we compare our approach to related work, before presenting the model and training procedure in detail in Sec. \ref{sec:our_model}. We evaluate our approach on different datasets in Sec \ref{sec:experiments}, showing promising performance and a high degree of generality for data types ranging from 2D points to audio snippets and images, and discuss these results with conclusions for future work in Sec. \ref{sec:conclusions}.

\section{Related Work}
\label{sec:related_work}

Learning to cluster based on neural networks has been approached mostly as a supervised learning problem to extract embeddings for a subsequent off-line clustering phase. The core of all deep metric embedding models is the choice of the loss function. Motivated by the fact that the softmax-cross entropy loss function has been designed as a classification loss and is not suitable for the clustering problem per se, \emph{Chopra et al.} \cite{DBLP:conf/cvpr/ChopraHL05} developed a ``Siamese'' architecture, where the loss function is optimized in a way to generate similar features for objects belonging to the same class, and dissimilar features for objects belonging to different classes. A closely related loss function called ``triplet loss'' has been used by \emph{Schroff et al.} \cite{schroff2015facenet} to get state-of-the-art accuracy in face detection. The main difference from the Siamese architecture is that in the latter case, the network sees same and different class objects with every example. It is then optimized to jointly learn their feature representation. A problem of both approaches is that they are typically difficult to train compared to a standard cross entropy loss.

\emph{Song et al.} \cite{DBLP:conf/cvpr/SongXJS16} developed an algorithm for taking full advantage of all the information available in training batches. They later refined the work \cite{DBLP:conf/cvpr/SongJR017} by proposing a new metric learning scheme based on structured prediction, which is designed to optimize a clustering quality metric (normalized mutual information \cite{mcdaid2011normalized}). Even better results were achieved by \emph{Wong et al.} \cite{DBLP:conf/iccv/WangZWLL17}, where the authors proposed a novel angular loss, and achieved state-of-the-art results on the challenging real-world datasets \emph{Stanford Cars} \cite{DBLP:conf/iccvw/Krause13} and \emph{Caltech Birds} \cite{DBLP:journals/ijcv/BransonHWPB14}. On the other hand, \emph{Lukic et al.} \cite{lukic2016speaker} showed that for certain problems, a carefully chosen deep neural network can simply be trained with softmax-cross entropy loss and still achieve state-of-the-art performance in challenging problems like speaker clustering. Alternatively, \emph{Wu et al.} \cite{DBLP:conf/iccv/ManmathaWSK17} showed that state-of-the-art results can be achieved simply by using a traditional margin loss function and being careful on how sampling is performed during the creation of mini-batches.

On the other hand, attempts have been made recently that are more similar to ours in spirit, using deep neural networks only and performing clustering end-to-end \cite{DBLP:journals/corr/abs-1801-07648}. They are trained in a fully unsupervised fashion, hence solve a different task then the one we motivated above (that is inspired by speaker- or image clustering based on some human notion of similarity). Perhaps first to group objects together in an unsupervised deep learning based manner where \emph{Le et al.} \cite{DBLP:conf/icml/LeRMDCCDN12}, detecting high-level concepts like cats or humans. \emph{Xie et al.} \cite{xie2016unsupervised} used an autoencoder architecture to do clustering, but experimental evaluated it only simplistic datasets like \emph{MNIST}. CNN-based approaches followed, e.g. by \emph{Yang et al.} \cite{DBLP:conf/cvpr/YangPB16}, where clustering and feature representation are optimized together. \emph{Greff et al.} \cite{DBLP:conf/nips/GreffSS17} performed perceptual grouping (of pixels within an image into the objects constituting the complete image, hence a different task than ours) fully  unsupervised using a neural expectation maximization algorithm. Our work differs from above-mentioned works in several respects: it has no assumption on the type of data, and solves the different task of grouping whole input objects.

\section{A model for end-to-end clustering of arbitrary data}
\label{sec:our_model}

\begin{figure}[t!]
  \centering
  \includegraphics[width=1\columnwidth]{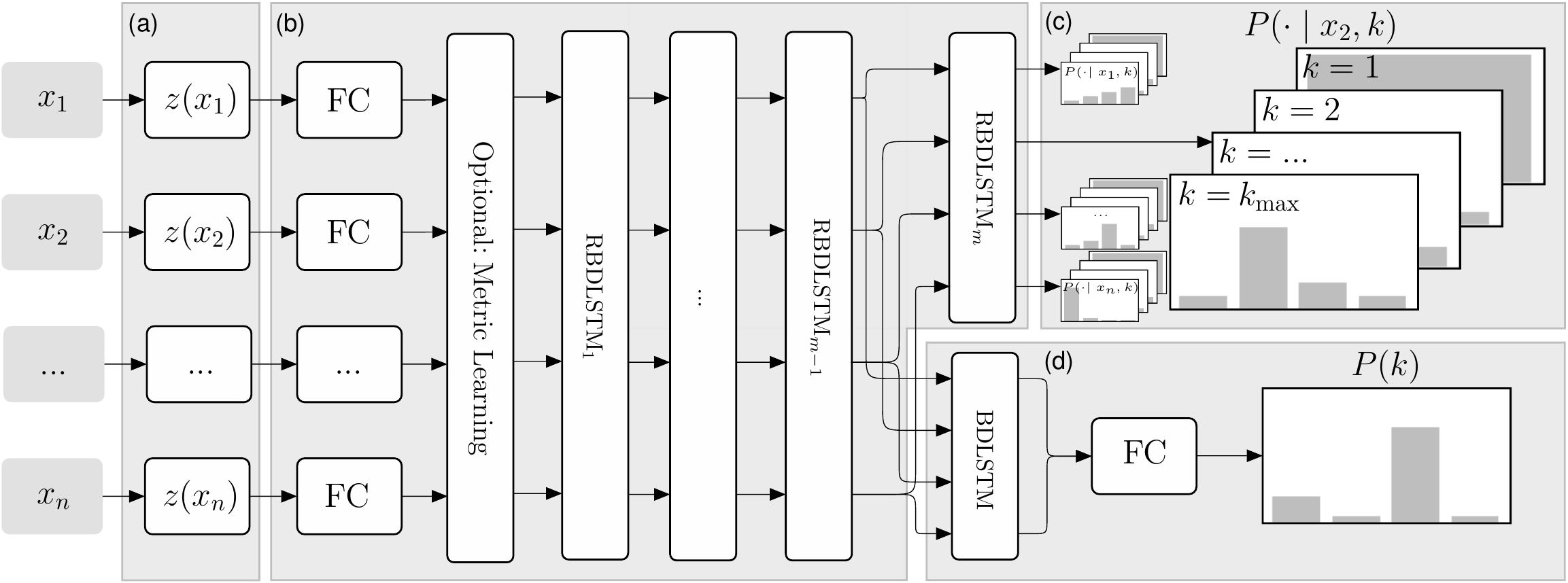}
  \caption{Our complete model, consisting of \mbox{(a)} the embedding network, \mbox{(b)} clustering network (including an optional metric learning part, see Sec. \ref{sec:metric_learning}), \mbox{(c)} cluster-assignment network and \mbox{(d)} cluster-count estimating network.}   
  \label{fig:clustering}
  \vspace{-0.6cm}
\end{figure}

Our method learns to cluster end-to-end purely ab initio, without the need to explicitly specify a notion of similarity, only providing the information whether two examples belong together. It uses as input $n \ge 2$ examples $x_i$, where $n$  may be different during training and application and constitutes the number of objects that can be clustered at a time, i.e. the maximum number of objects in a partition. The network's output is two-fold: a probability distribution $P(k)$ over the cluster count $1 \leq k \leq k_\mathrm{max}$; and probability distributions $P(\cdot\mid x_i,k)$ over all possible cluster indexes for each input example $x_i$ and for each $k$. 

\subsection{Network architecture}
The network architecture (see Fig. \ref{fig:clustering}) allows the flexible use of different input types, e.g. images, audio or 2D points. An input $x_i$ is first processed by an embedding network (a) that produces a lower-dimensional representation $z_i=z(x_i)$. The dimension of $z_i$ may vary depending on the data type. For example, 2D points do not require any embedding network. A fully connected layer (FC) with $\mathrm{LeakyReLU}$ activation at the beginning of the clustering network (b) is then used to bring all embeddings to the same size. This approach allows to use the identical subnetworks (b)--(d) and only change the subnet (a) for any data type. The goal of the subnet (b) is to compare each input $z(x_i)$ with all other $z(x_{j\ne i})$, in order to learn an abstract grouping which is then concretized into an estimation of the number of clusters (subnet (d)) and a cluster assignment (subnet (c)).

\begin{figure}[t!]
  \centering
  \includegraphics[width=0.5\columnwidth]{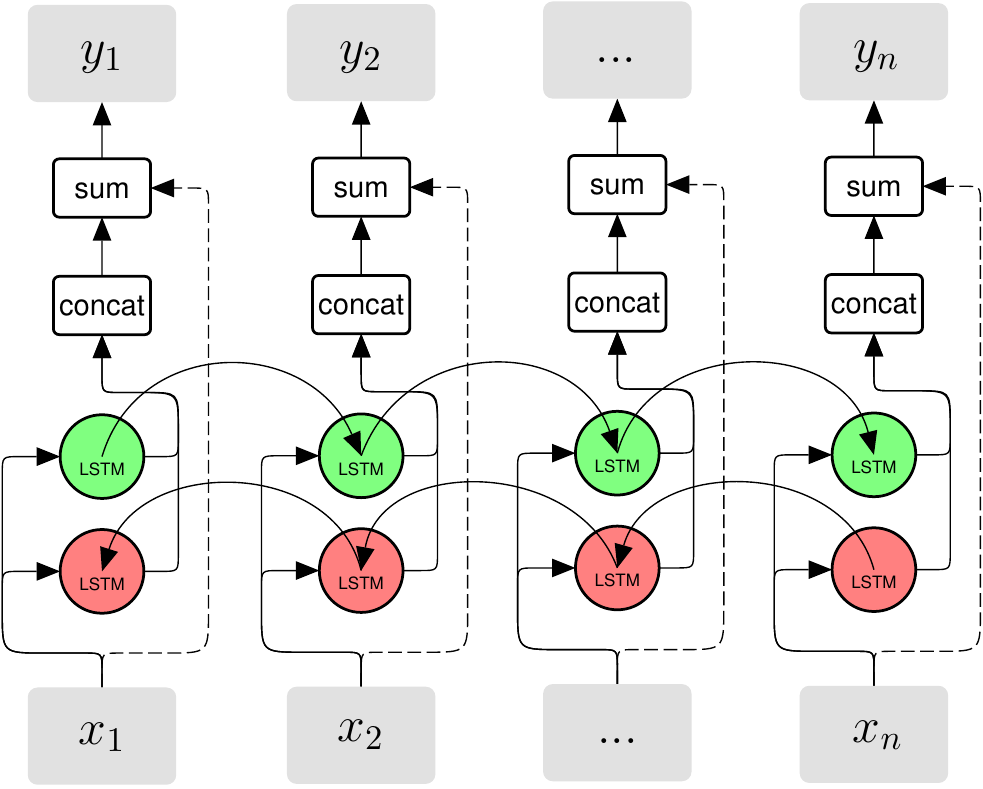}
  \caption{$\mathrm{RBDLSTM}$-layer: A $\mathrm{BDLSTM}$ with residual connections (dashed lines). The variables $x_i$ and $y_i$ are named independently from the notation in Fig. \ref{sec:our_model}.}
  \label{fig:rbdlstm}
  \vspace{-0.5cm}
\end{figure}

To be able to process a non-fixed number of examples $n$ as input, we use a recurrent neural network. Specifically, we use stacked residual bi-directional LSTM-layers ($\mathrm{RBDLSTM}$), which are similar to the cells described in \cite{wu2016google} and visualized in Fig. \ref{fig:rbdlstm}. The residual connections allow a much more effective gradient flow during training \cite{he2016deep} and avoid vanishing gradients. Additionally, the network can learn to use or bypass certain layers using the residual connections, thus reducing the architectural decision on the number of recurrent layers to the simpler one of finding a reasonable upper bound.

The first of overall two outputs is modeled by the cluster assignment network \mbox{(c)}. It contains a $\mathrm{softmax}$-layer to produce $P(\ell\mid x_i,k)$, which assigns a cluster index $\ell$ to each input $x_i$, given $k$ clusters (i.e., we get a distribution over possible cluster assignments for each input and every possible number of clusters). The second output, produced by the cluster-count estimating network \mbox{(d)}, is built from another $\mathrm{BDLSTM}$-layer. Due to the bi-directionality of the network, we concatenate its first and the last output vector into a fully connected layer of twice as many units using again $\mathrm{LeakyReLUs}$. The subsequent $\mathrm{softmax}$-activation finally models the distribution $P(k)$ for $1\leq k\leq k_\mathrm{max}$. The next subsection shows how this neural network learns to approximate these two complicated probability distributions \cite{Lee2017OnTA} purely from pairwise constraints on data that is completely separate from any dataset to be clustered. No labels for clustering are needed.

\subsection{Training and loss}
In order to define a suitable loss-function, we first define an approximation (assuming independence) of the probability that $x_i$ and $x_j$ are assigned to the same cluster for a given $k$ as
\begin{align*}
  P_{ij}(k)=\sum_{\ell=1}^{k} P(\ell\mid x_i,k)P(\ell\mid x_j,k).
\end{align*}
By marginalizing over $k$, we obtain $P_{ij}$, the probability that $x_i$ and $x_j$ belong to the same cluster:
\begin{align*}
   P_{ij}=\sum_{k=1}^{k_{\mathrm{max}}} P(k) \sum_{\ell=1}^{k} P(\ell\mid x_i,k)P(\ell\mid x_j,k).
\end{align*}
Let $y_{ij}=1$ if $x_i$ and $x_j$ are from the same cluster (e.g., have the same group label) and $0$ otherwise. The loss component for \emph{cluster assignments}, $L_\mathrm{ca}$, is then given by the weighted binary cross entropy as
\begin{align*}
  L_{\mathrm{ca}}=\frac{-2}{n(n-1)}\sum_{i<j}{\left(\varphi_1 y_{ij} \log(P_{ij})+\varphi_2 (1-y_{ij}) \log(1-P_{ij})\right)}
\end{align*}
with weights $\varphi_1$ and $\varphi_2$. The idea behind the weighting is to account for the imbalance in the data due to there being more dissimilar than similar pairs $(x_i,x_j)$ as the number of clusters in the mini batch exceeds $2$. Hence, the weighting is computed using $\varphi_1=c\sqrt{1-\varphi}$ and $\varphi_2=c\sqrt{\varphi}$, with $\varphi$ being the expected value of $y_{ij}$ (i.e., the a priori probability of any two samples in a mini batch coming from the same cluster), and $c$ a normalization factor so that $\varphi_1 + \varphi_2 = 2$. The value $\varphi$ is computed over all possible cluster counts for a fixed input example count $n$, as during training, the cluster count is randomly chosen for each mini batch according to a uniform distribution. The weighting of the cross entropy given by $\varphi$ is then used to make sure that the network does not converge to a sub-optimal and trivial minimum. Intuitively, we thus account for permutations in the sequence of examples by checking rather for pairwise correctness (probability of same/different cluster) than specific indices.

The second loss term, $L_\mathrm{cc}$, penalizes a wrong \emph{number of clusters} and is given by the categorical cross entropy of $P(k)$ for the true number of clusters $k$ in the current mini batch:
\begin{align*}
  L_\mathrm{cc}&= -\log(P(k)).
\end{align*}

The complete loss is given by $L_{\mathrm{tot}}=L_{\mathrm{cc}}+ \lambda L_{\mathrm{ca}}$. During training, we prepare each mini batch with $N$ sets of $n$ input examples, each set with $k=1\ldots k_\mathrm{max}$ clusters chosen uniformly. Note that this training procedure requires only the knowledge of $y_{ij}$ and is thus also possible for weakly labeled data. All input examples are randomly shuffled for training and testing to avoid that the network learns a bias w.r.t. the input order. To demonstrate that the network really learns an intra-class distance and not just classifies objects of a fixed set of classes, it is applied on totally different clusters at evaluation time than seen during training.

\subsection{Implicit vs. explicit distance learning}
\label{sec:metric_learning}

To elucidate the importance and validity of the implicit learning of distances in our subnetwork (b), we also provide a modified version of our network architecture for comparison, in which the calculation of the distances is done explicitly. Therefore, we add an extra component to the network before the RBDLSTM layers, as can be seen in Figure \ref{fig:clustering}: the optional metric learning block receives the fixed-size embeddings from the fully connected layer after the embedding network (a) as input and outputs the pairwise distances of the data points. The recurrent layers in block (b) then subsequently cluster the data points based on this pairwise distance information \cite{chin2010novel,arias2011clustering} provided by the metric learning block. 

We construct a novel metric learning block inspired by the work of \emph{Xing et al.} \cite{xing2003distance}. In contrast to their work, we optimize it end-to-end with backpropagation. This has been proposed in \cite{schwenker2001three} for classification alone; we do it here for a clustering task, for the whole covariance matrix, and jointly with the rest of our network. We construct the non-symmetric, non-negative dissimilarity measure $d^2_A$ between two data points $x_i$ and $x_j$ as
\begin{align*}
  d^2_A(x_i, x_j) =  (x_i-x_j)^{T}A(x_i-x_j)
  \label{eq:dist}
\end{align*}
and let the neural network training optimize $A$ through $L_{\mathrm{tot}}$ without intermediate losses. The matrix $A$ as used in $d^2_A$ can be thought of as a trainable distance metric. In every training step, it is projected into the space of positive semidefinite matrices.

\section{Experimental results}
\label{sec:experiments}

To assess the quality of our model, we perform clustering on three different datasets: for a proof of concept, we test on a set of generated \emph{2D points} with a high variety of shapes, coming from different distributions. For speaker clustering, we use the \emph{TIMIT} \cite{timit} corpus, a dataset of studio-quality speech recordings frequently used for pure speaker clustering in related work. For image clustering, we test on the \emph{COIL-100} \cite{nayar1996columbia} dataset, a collection of different isolated objects in various orientations. To compare to related work, we measure the performance with the standard evaluation scores misclassification rate (MR) \cite{liu2003online} and normalized mutual information (NMI) \cite{mcdaid2011normalized}. Architecturally, we choose $m=14$ BDLSTM layers and $288$ units in the FC layer of subnetwork (b), $128$ units for the BDLSTM in subnetwork (d), and $\alpha=0.3$ for all $\mathrm{LeakyReLUs}$ in the experiments below. All hyperparameters where chosen based on preliminary experiments to achieve reasonable performance, but not tested nor tweaked extensively. The code and further material and experiments are available online\footnote{See \url{https://github.com/kutoga/learning2cluster}.}.

\begin{figure}[t!]
    \centering
    \begin{subfigure}[t]{0.31\columnwidth} 
        \centering
    	\vspace{-68.92px}
        \includegraphics[width=0.909\columnwidth]{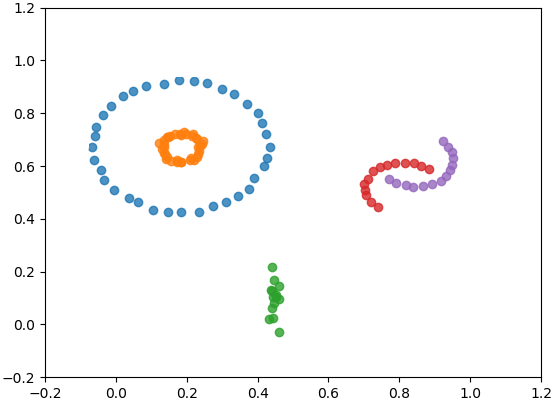}
        \caption{}
        \label{fig:output_clustering_1}
    \end{subfigure}
    \begin{subfigure}[t]{0.31\columnwidth}
        \centering
        \includegraphics[width=0.90\columnwidth]{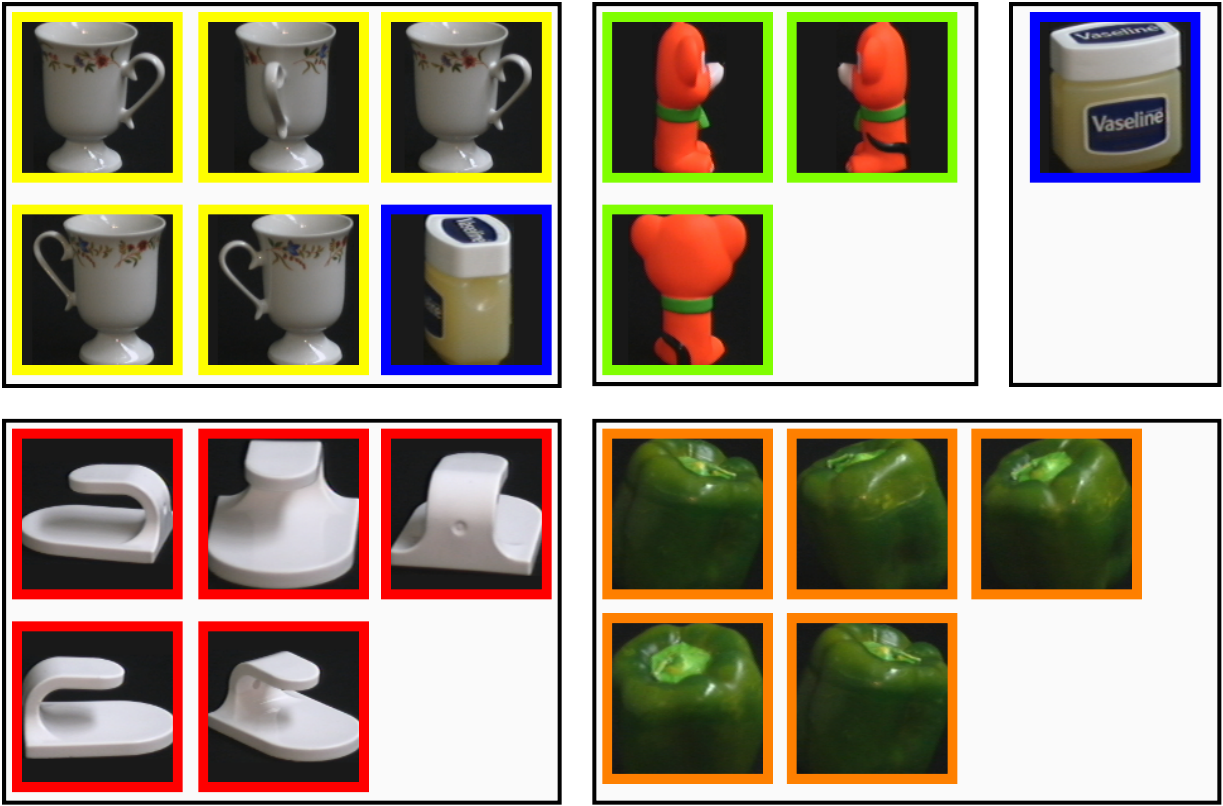}
        \caption{}
        \label{fig:output_clustering_2}
    \end{subfigure}
    \begin{subfigure}[t]{0.31\columnwidth}
        \centering
        \includegraphics[width=0.90\columnwidth]{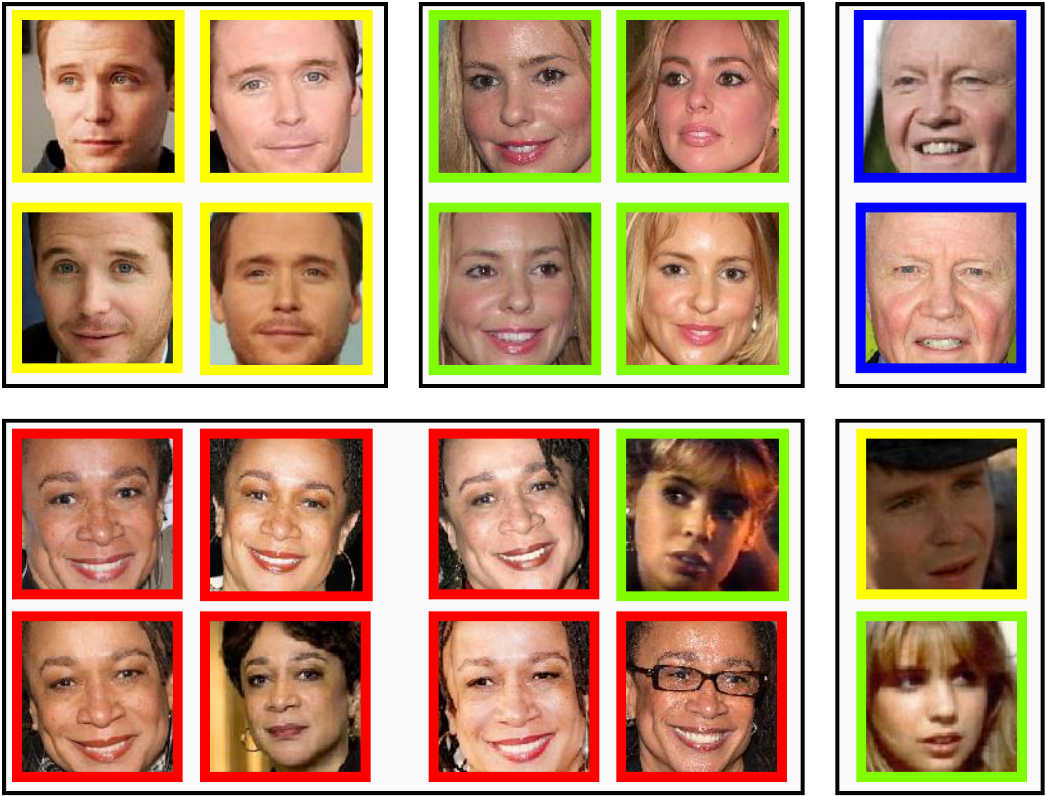}
        \caption{}
        \label{fig:output_clustering_3}
    \end{subfigure}
  \caption{Clustering results for (a) 2D point data, (b) COIL-100 objects, and (c) faces from FaceScrub (for illustrative purposes). The color of points / colored borders of images depict true cluster membership.}
  \label{fig:output_2d_point_clustering}
  \vspace{-0.5cm}
\end{figure}

We set $k_\mathrm{max}=5$ and $\lambda=5$ for all experiments. For the 2D point data, we use $n=72$ inputs and a batch-size of $N=200$ (We used the batch size of $N=50$ for metric learning with 2D points). For TIMIT, the network input consists of $n=20$ audio snippets with a length of $1.28$ seconds, encoded as mel-spectrograms with $128\times 128$ pixels (identical to \cite{lukic2017speaker}). For COIL-100, we use $n=20$ inputs with a dimension of $128\times 128\times 3$. For TIMIT and \mbox{COIL-100}, a simple CNN with 3 conv/max-pooling layers is used as subnetwork (a). For TIMIT, we use $430$ of the $630$ available speakers for training (and $100$ of the remaining ones each for validation and evaluation). For \mbox{COIL-100}, we train on $80$ of the $100$ classes ($10$ for validation, $10$ for evaluation). For all runs, we optimize using Adadelta \cite{zeiler2012adadelta} with a learning rate of $5.0$. Example clusterings are shown in Fig. \ref{fig:output_2d_point_clustering}. For all configurations, the used hardware set the limit on parameter values: we used the maximum possible batch size and values for $n$ and $k_\mathrm{max}$ that allow reasonable training times. However, values of $n\ge 1000$ where tested and lead to a large decrease in model accuracy. This is a major issue for future work.

\begin{table}[t!]
  \scriptsize
  \centering
  \caption{$\mathrm{NMI} \in [0,1]$ and $\mathrm{MR} \in [0,1]$ averaged over $300$ evaluations of a trained network. We abbreviate our ``learning to cluster'' method as ``L2C''.}
  \label{tbl:results}
  \begin{tabular}{|l|l|l|l|l|l|l|}
  \hline
             & \multicolumn{2}{l|}{\textbf{2D Points (self generated)}} & \multicolumn{2}{l|}{\textbf{TIMIT}} & \multicolumn{2}{l|}{\textbf{COIL-100}} \\ \hline
             & \textbf{MR}        & \textbf{NMI}       & \textbf{MR}      & \textbf{NMI}     & \textbf{MR}       & \textbf{NMI}       \\ \hline
  L2C ($=$our method) & $0.004$ & $0.993$ & $0.060$ & $0.928$ & $0.116$ & $0.867$ \\ \hline
  L2C + Euclidean & $0.177$ & $0.730$ & $0.093$  & $0.883$ & $0.123$  & $0.884$  \\ \hline
  L2C + Mahalanobis & $0.185$ & $0.725$ & $0.104$ & $0.882$ & $0.093$ & $0.890$ \\ \hline
  L2C + Metric Learning & $0.165$ & $0.740$ & $0.101$ & $0.880$ & $0.100$ & $0.880$ \\ \hline
  Random cluster assignment & $0.485$ & $0.232$ & $0.435$ & $0.346$ & $0.435$ & $0.346$ \\   \hline
  Baselines (related work) &
  \multicolumn{2}{l|}{
  \makecell[l]{
    k-Means: $\mathrm{MR}=0.178$, $\mathrm{NMI}=0.796$\\
    DBSCAN: $\mathrm{MR}=0.265$, $\mathrm{NMI}=0.676$\\
  }
  }
  &
  \multicolumn{2}{l|}{
  \makecell{
  \cite{lukic2017speaker}: $\mathrm{MR}=0$
  }}
  &
  \multicolumn{2}{l|}{
  \makecell{
  \cite{DBLP:conf/cvpr/YangPB16}: $\mathrm{NMI}=0.985$
  }}
  \\ \hline
  \end{tabular}
  \vspace{-0.5cm}
\end{table}

The results on 2D data as presented in Fig. \ref{fig:output_clustering_1} demonstrate that our method is able to learn specific and diverse characteristics of intuitive groupings. This is superior to any single traditional method, which only detects a certain class of cluster structure (e.g., defined by distance from a central point). Although \cite{lukic2017speaker} reach moderately better scores for the speaker clustering task and \cite{DBLP:conf/cvpr/YangPB16} reach a superior $\mathrm{NMI}$ for \mbox{COIL-100}, our method finds reasonable clusterings, is more flexible through end-to-end training and is not tuned to a specific kind of data. Hence, we assume, backed by the additional experiments to be found online, that our model works well also for other data types and datasets, given a suitable embedding network. \mbox{Tab. \ref{tbl:results}} gives the numerical results for said datasets in the row called ``L2C'' without using the explicit metric learning block. Extensive preliminary experiments on other public datasets like e.g. FaceScrub \cite{ng2014data} confirm these results: learning to cluster reaches promising performance while not yet being on par with tailor-made state-of-the-art approaches. 

We compare the performance of our implicit distance metric learning method to versions enhanced by  different explicit schemes for pairwise similarity computation prior to clustering. Specifically, three implementations of the optional metric learning block in subnetwork (b) are evaluated: using a fixed  diagonal matrix $A$ (resembling the Euclidean distance), training a diagonal $A$ (resembling Mahalanobis distance), and learning the entire coefficients of the distance matrix $A$. Since we argue above that our approach combines \emph{implicit} deep metric embedding with clustering in an end-to-end architecture, one would not expect that adding \emph{explicit} metric computation changes the results by a large extend. This assumption is largely confirmed by the results in the ``L2C$+$\dots'' rows in \mbox{Tab. \ref{tbl:results}}: for COIL-100, Euclidean gives slightly worse, and the other two slightly better results than L2C alone; for TIMIT, all results are worse but still reasonable. We attribute the considerable performance drop on 2D points using all three explicit schemes to the fact that in this case much more instances are to be compared with each other (as each instance is smaller than e.g. an image, $n$ is larger). This might have needed further adaptations like e.g. larger batch sizes (reduced here to $N=50$ for computational reasons) and longer training times.

\section{Discussion and conclusions}
\label{sec:conclusions}

We have presented a novel approach to learn neural models that directly output a probabilistic clustering on previously unseen groups of data; this includes a solution to the problem of outputting similar but unspecific ``labels'' for similar objects of unseen ``classes''. A trained model is able to cluster different data types with promising results. This is a complete end-to-end approach to clustering that learns both the relevant features and the ``algorithm'' by which to produce the clustering itself. It outputs probabilities for cluster membership of all inputs as well as the number of clusters in test data. The learning phase only requires pairwise labels between examples from a separate training set, and no explicit similarity measure needs to be provided. This is especially useful for high-dimensional, perceptual data like images and audio, where similarity is usually semantically defined by humans. Our experiments confirm that our algorithm is able to implicitly learn a metric and directly use it for the included clustering. This is similar in spirit to the very recent work of \emph{Hsu et al.} \cite{hsu2018learning}, but does not need and optimization on the test (clustering) set and finds $k$ autonomously. It is a novel approach to \emph{learn to cluster}, introducing a novel architecture and loss design.

We observe that the clustering accuracy depends on the availability of a large number of different classes during training. We attribute this to the fact that the network needs to learn intra-class distances, a task inherently more difficult than just to distinguish between objects of a fixed amount of classes like in classification problems. We understand the presented work as an early investigation into the new paradigm of learning to cluster by perceptual similarity specified through examples. It is inspired by our work on speaker clustering with deep neural networks, where we increasingly observe the need to go beyond surrogate tasks for learning, training end-to-end specifically for clustering to close a performance leak. While this works satisfactory for initial results, points for improvement revolve around scaling the approach to practical applicability, which foremost means to get rid of the dependency on $n$ for the partition size.

The number $n$ of input examples to assess simultaneously is very relevant in practice: if an input data set has thousands of examples, incoherent single clusterings of subsets of $n$ points would be required to be merged to produce a clustering of the whole dataset based on our model. As the (RBD)LSTM layers responsible for assessing points simultaneously in principle have a long, but still local (short-term) horizon, they are not apt to grasp similarities of thousands of objects. Several ideas exist to change the architecture, including to replace recurrent layers with temporal convolutions, or using our approach to seed some sort of differentiable K-means or EM layer on top of it. Preliminary results on this exist. Increasing $n$ is a prerequisite to also increase the maximum number of clusters $k$, as $k\ll n$. For practical applicability, $k$ needs to be increased by an order of magnitude; we plan to do this in the future. This might open up novel applications of our model in the area of transfer learning and domain adaptation.

\paragraph{\textbf{Acknowledgements:} We thank the anonymous reviewers for helpful feedback.}

\bibliography{annpr2018-2}
\bibliographystyle{splncs04}

\end{document}